\documentclass[10pt,twocolumn,letterpaper]{article}

\usepackage{cvpr}
\usepackage{times}
\usepackage{epsfig}
\usepackage{graphicx}
\usepackage{tabularx}
\usepackage{amsmath}
\usepackage{amssymb}
\usepackage{bm}
\usepackage{bbm}
\usepackage{algorithm}
\usepackage{algorithmic}
\usepackage{booktabs}
\usepackage{multirow}
\usepackage{tabularx}
\usepackage{xcolor}
\usepackage{subfig}
\usepackage{authblk}
\colorlet{GRAY}{gray}

\makeatletter
\def\ps@myheadings{%
	\let\@oddfoot\@empty\let\@evenfoot\@empty
	\def\@evenhead{\thepage\hfil\slshape\leftmark}%
	\def\@oddhead{{\slshape\rightmark}\hfil\thepage}%
	\let\@mkboth\@gobbletwo
	\let\sectionmark\@gobble
	\let\subsectionmark\@gobble
}
\if@titlepage
\renewcommand\maketitle{\begin{titlepage}%
		\let\footnotesize\small
		\let\footnoterule\relax
		\let \footnote \thanks
		\null\vfil
		\vskip 60\p@
		\begin{center}%
			{\LARGE \@title \par}%
			\vskip 3em%
			{\large
				\lineskip .75em%
				\begin{tabular}[t]{c}%
					\@author
				\end{tabular}\par}%
			\vskip 1.5em%
			{\large \@date \par}
		\end{center}\par
		\@thanks
		\vfil\null
	\end{titlepage}%
	\setcounter{footnote}{0}%
}
\else
\renewcommand\maketitle{\par
	\begingroup
	\renewcommand\thefootnote{\@fnsymbol\c@footnote}%
	\def\@makefnmark{\rlap{\@textsuperscript{\normalfont\@thefnmark}}}%
	\long\def\@makefntext##1{\parindent 1em\noindent
		\hb@xt@1.8em{%
			\hss\@textsuperscript{\normalfont\@thefnmark}}##1}%
	\if@twocolumn
	\ifnum \col@number=\@ne
	\@maketitle
	\else
	\twocolumn[\@maketitle]%
	\fi
	\else
	\newpage
	\global\@topnum\z@   
	\@maketitle
	\fi
	\thispagestyle{plain}\@thanks
	\endgroup
	\setcounter{footnote}{0}%
}
\makeatother

\usepackage{array}
\newcolumntype{?}{!{\vrule width 1pt}}
\newcolumntype{C}[1]{>{\centering\arraybackslash\hspace{0pt}}p{#1}}

\graphicspath{{images/}}

\usepackage{color}

\newcommand{\comment}[1]{}

\newif\ifdraft
\drafttrue

\ifdraft
 
 \newcommand{\BT}[1]{{\color{blue}BT: {\bf #1}}}
 
 \newcommand{\FB}[1]{{\color{violet}FB: {\bf #1}}}
 
 \newcommand{\MP}[1]{{\color{red}MP: {\bf #1}}}
\else
 
 \newcommand{\BT}[1]{}
 
 \newcommand{\FB}[1]{}
 
 \newcommand{\MP}[1]{}
\fi

\newcommand{\by}[0]{\mathbf{y}}

\newcommand{\bI}{\mathbf{I}}
\newcommand{\bG}{\mathbf{G}}

\newcommand{\bp}{\mathbf{p}}

\newcommand{\bv}{\mathbf{v}}

\newcommand{\real}{\mathbb{R}}

\newcolumntype{M}[1]{>{\centering\arraybackslash}m{#1}}
\newcolumntype{R}[1]{>{\raggedleft\arraybackslash}m{#1}}
\newcolumntype{P}[1]{>{\centering\arraybackslash}p{#1}}

\definecolor{mygray}{gray}{0.2}

\usepackage[font=small,labelfont=bf]{caption}

\usepackage[pagebackref=true,breaklinks=true,letterpaper=true,colorlinks,bookmarks=false]{hyperref}

\def\cvprPaperID{2754} 

\cvprfinalcopy
\ifcvprfinal\fi
\begin{document}

\title{H+O: Unified Egocentric Recognition of 3D Hand-Object Poses and Interactions}


\author{Bugra Tekin$^1$ \quad\quad\quad\quad Federica Bogo$^1$ \quad\quad\quad\quad Marc Pollefeys$^{1,2}$ \\
		$^1$ Microsoft \quad\quad\quad\quad $^2$ ETH Z\"{u}rich
}

\pagenumbering{gobble}


\maketitle

\begin{abstract}

We present a unified framework for understanding 3D hand and object interactions in raw image sequences from egocentric RGB cameras. Given a single RGB image, our model jointly estimates the 3D hand and object poses, models their interactions, and recognizes the object and action classes with a single feed-forward pass through a neural network. We propose a single architecture that does not rely on external detection algorithms but rather is trained end-to-end on single images. We further merge and propagate information in the temporal domain to infer interactions between hand and object trajectories and recognize actions. The complete model takes as input a sequence of frames and outputs per-frame 3D hand and object pose predictions along with the estimates of object and action categories for the entire sequence. We demonstrate state-of-the-art performance of our algorithm even in comparison to the approaches that work on depth data and ground-truth annotations.

\end{abstract} 


\section{Introduction}

\begin{figure}[t]
	\begin{center}
		\includegraphics[width=\linewidth]{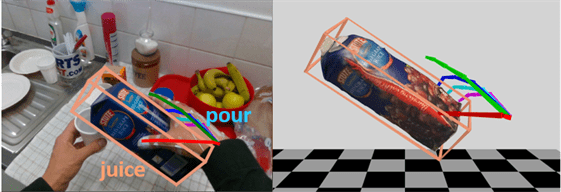}
	\end{center}
	\caption{Unified reasoning on first-person views. Our method takes as input color images and produces a comprehensive egocentric scene interpretation. We estimate simultaneously 3D hand and object poses (shown as skeletons and 3D bounding boxes), object class (\eg juice bottle) and action category (\eg pouring).}
	\label{fig:teaser}
\end{figure}

Human behavior can be characterized by the individual actions a person takes in interaction with the surrounding objects and the environment. A significant amount of research has focused on visual understanding of humans~\cite{Choutas18,Iqbal18,Mueller18,Oberweger15,Sridhar15,Tekin17,Ye18,Zimmermann17} and objects~\cite{Brachmann16,Kehl17,Tejani14,Tekin18}, in isolation from each other. However, the problem of jointly understanding humans and objects,  although crucial for a semantically meaningful interpretation of the visual scene, has received far less attention. In this work, we propose, for the first time, a unified method to jointly recognize 3D hand and object poses, and their interactions from egocentric monocular color images. Our method jointly estimates the hand and object poses in 3D, models their interactions and recognizes the object and activity classes. An example result is shown in  Fig.~\ref{fig:teaser}. Our unified  framework is highly relevant for augmented and virtual reality~\cite{Surie07}, fine-grained recognition of people's actions, robotics and telepresence.

Capturing hands in action while taking into account the objects in contact is an extremely challenging problem. Jointly reasoning about hands and objects from moving, egocentric cameras is even more challenging as this would require understanding of the complex and often subtle interactions that take place in cluttered real-world scenes where the hand is often occluded by the object or the viewpoint. Recent research in computer vision has successfully addressed some of the challenges in joint understanding of hands and objects for depth and multi-camera input. Sridhar et al.~\cite{Sridhar16} have demonstrated that accounting jointly for hands and objects helps predict the 3D pose more accurately than models that ignore interaction. Pioneering works by~\cite{Hamer09,Mueller17,Oikonomidis11,Tzionas16} have proposed ways to model hand-object interactions to increase robustness and accuracy in recovering the hand motion.

Most of these works, however, are limited by the following  factors: Firstly, they either rely on active depth sensors or 
multi-camera systems. Depth sensors are power hungry and less prevalent than regular RGB cameras. On the other hand, multi-camera 
systems are impractical due to the cost and effort in setting up a calibrated and synchronous system of sensors. Secondly, 
they do not reason about the action the subject is performing. While estimating the 3D hand pose is crucial for many 
applications in robotics and graphics, the sole knowledge of the pose lacks semantic meaning about the actions of the subject. 
Thirdly, they focus mostly on only capturing hand motion without recovering the object pose in 3D~\cite{Hamer09,Mueller17}, and, therefore, lack 
environmental understanding.\comment{mention than egocentric is more difficult. I think you should also mention your CVPR18,
and state that this work extends the approach in a non-trivial way.}\comment{\BT{done}}

Our method aims at tackling all these issues. To this end, we propose an approach for predicting simultaneously 3D hand and 
object poses, object classes and action categories through a novel data-driven architecture. Our model jointly produces 3D hand 
and object pose estimates, action and object classes from a single image and requires neither external region proposals nor pre-computed 
detections\comment{\FB{Isn't it common to all YOLO architectures?}\BT{Indeed, but still it's an advantage over other pose estimation methods}}. Our method further models the temporal nature of 3D hand and object motions to recognize actions and infer interactions.

Our contributions can be summarized as follows:
\vspace{-1mm}
\begin{itemize}
    \item We propose a unified framework for recognizing 3D hand and object interactions
	by simultaneously solving four tasks in a feed-forward pass through a neural network:
	3D hand pose estimation, object pose estimation, object recognition 
	and activity classification. Our method operates on monocular color images and relies on joint features that are shared among all the tasks.
	\vspace{-1mm}
	\item We introduce a novel single-shot neural network framework that jointly solves for the 3D articulated and rigid pose estimation problems within the same architecture. Our scheme relies on a common output representation for both hands and objects that parametrizes their pose with 3D control points. Our network directly predicts the control points in 3D rather than in 2D, in contrast to the common single-shot neural network paradigm~\cite{Redmon17,Tekin18}, dispenses with the need to solve a 2D-to-3D correspondence problem~\cite{Lepetit09} and yields large improvements in accuracy. 
	 \vspace{-1mm}
	\item We present a temporal model to merge and propagate information in the temporal domain, explicitly model
	interactions and infer relations between hand and objects, directly in 3D. 
\end{itemize}

In  Section~\ref{sec:results},  we  show  quantitatively  that  these contributions allow us to achieve better overall performance in targeted tasks, while running at real-time speed and not requiring detailed 3D hand and object models.  Our approach,  which we call \emph{Hand + Object (H+O)}, achieves state-of-the-art results on challenging sequences, and outperforms existing approaches that rely on the ground-truth pose annotations and depth data.


\section{Related Work}
\label{sec:related}

We now review existing work on 3D hand and object pose estimation -- both jointly and in isolation -- and action recognition, with a focus on egocentric scenarios.

\paragraph{Hands and Objects.}
Many approaches in the literature tackle the problem of estimating either hand or object pose in isolation.

Brachman et al.~\cite{Brachmann16} recover 6D object pose from single RGB images using a multi-stage approach, based on regression forests. More recent approaches~\cite{Kehl17,Rad17} rely on Convolutional Neural Networks (CNNs). BB8~\cite{Rad17} uses CNNs to roughly segment the object and then predict the 2D locations of the object's 3D bounding box, projected in image space. 6D pose is then computed from these estimates via PnP~\cite{Lepetit09}. SSD-6D~\cite{Kehl17} predicts 2D bounding boxes together with an estimate of the object pose. These methods need a detailed textured 3D object model as input, and require a further pose refinement step to improve accuracy. Tekin et al.~\cite{Tekin18} overcome these limitations by introducing a single-shot architecture which predicts 2D projections of the object's 3D bounding box in a single forward pass, at real-time speed. All these approaches do not address the problem of estimating object pose in hand-object interaction scenarios, where objects might be largely occluded.

3D hand pose and shape estimation in egocentric views has started receiving attention recently~\cite{Choi17,Hasson19,Mueller17,Rogez15cvpr,Ye18,Yuan18,Yuan17}.
The problem is challenging with respect to third-person scenarios~\cite{Spurr18,Zimmermann17}, due to self-occlusions and the limited amount of training data available~\cite{Mueller17,Oberweger16,Rogez15cvpr,Ye18}. Mueller et al.~\cite{Mueller17} train CNNs on synthetic data and combine them with a generative hand model to track hands interacting with objects from egocentric RGB-D videos. This hybrid approach has then been extended to work with RGB videos~\cite{Mueller18}. Iqbal et al.~\cite{Iqbal18} estimate 3D hand pose from single RGB images, from both first- and third-person views, regressing 2.5D heatmaps via CNNs. These methods focus on hand pose estimation and try to be robust in the presence of objects, but do not attempt to model hand and objects together. 

While reasoning about hands in action, object interactions can be exploited as additional constraints~\cite{Oikonomidis11,Ren10,Rogez15cvpr,Romero10}. By observing that different object shapes induce different hand grasps, the approaches in~\cite{Choi17,Rogez15cvpr} discriminatively estimate 3D hand pose from depth input. Model-based approaches~\cite{Romero17} have also been proposed to jointly estimate hand and object parameters at a finer level of detail. However, most approaches focus on third-view scenarios, taking depth as input~\cite{Oikonomidis11,Panteleris15,Tsoli18,Tzionas16}.

To our knowledge, no approach in the literature jointly estimates 3D hand and object pose from RGB video only.

\paragraph{Action Recognition.}

While action recognition is a long-standing problem in computer vision~\cite{Bobick01,Dollar05,Gkioxari18,Jhuang07,Laptev03,Niebles06,Wong07}, first-person action recognition started to emerge as an active field only recently, largely due to the advent of consumer-level wearable sensors and large egocentric datasets~\cite{Damen18,Goyal2017,Pirsiavash12,Rogez15iccv}. First-person views bring upon unique challenges to action recognition due to fast camera motion, large occlusions and background clutter~\cite{Li15}. 

Early approaches for egocentric action recognition rely on motion cues~\cite{Kitani11,Ren13,Ryoo15}. In particular,~\cite{Kitani11} uses optical flow-based global motion descriptors to categorize ``ego-actions'' across sports genres. ~\cite{Poleg16} feeds sparse optical flow to 3D CNNs to index egocentric videos. Motion and appearance cues are used in conjunction with depth in~\cite{Tang17}. In addition to motion, features based on gaze information~\cite{Fathi12}, head motion~\cite{Li15}, and, recently, CNN-learned features~\cite{Ma16} have been proposed by a number of studies. 
Another line of work has focused specifically on hand and object cues for first person action recognition~\cite{Fathi11iccv,Fouhey18,Kalogeiton17,Sigurdsson18,Sundaram09}. Pirsiavash and Ramanan~\cite{Pirsiavash12} explore active object detection as an auxiliary task for activity recognition. Koppula et al.~\cite{Koppula13} learn a model of object affordances to understand activities from RGB-D videos. EgoNet~\cite{Berta17} detects ``action objects'', \ie objects linked to visual or tactile interactions, from first-person RGB-D images.
Fathi et al.~\cite{Fathi11iccv,Fathi11cvpr} use motion cues to segment hands and objects, and then extract features from these foreground regions.
All these approaches, however, focus on 2D without explicitly modeling hand-object interactions in 3D. Recently, Garcia-Hernando et al.~\cite{GarciaHernando18} demonstrate that 3D heuristics are beneficial to first person action recognition. However, they work on depth input and rely on ground-truth object poses.

Similarly to us, Cai et al.~\cite{Cai16} propose a structured approach where grasp types, object attributes and their contextual relationships are analyzed together. However, their single-image framework does not consider the temporal dimension. Object Relation Network~\cite{Baradel18} models contextual relationships between detected semantic object instances, through space and time. All these approaches aim at understanding the scene only in 2D. Here, we model more complex hand and object attributes in 3D, as well as their temporal interaction.


\section{Method}
\label{sec:method}

Our goal is to construct comprehensive interpretations of egocentric scenes from raw image sequences to understand human activities. To this end, we propose a unified framework to jointly estimate 3D hand and object poses and recognize object and action classes.

\begin{figure*}
	\begin{center}
		\includegraphics[width=\linewidth]{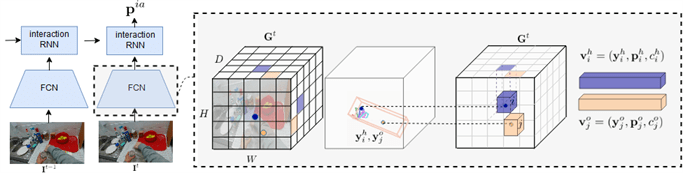}
	\end{center}
	\vspace{-3mm}
	\footnotesize \hspace{2cm} \textbf{(a)} \hspace{3.8cm} \textbf{(b)} \hspace{2.4cm} \textbf{(c)} \hspace{3cm} \textbf{(d)} \hspace{2.5cm} \textbf{(e)}\hspace{1cm}  \\
	\vspace{-5mm}
	\caption{Overview of our \emph{Hand+Object} approach. \textbf{(a)} The proposed network architecture. Each frame $\bI^t$ is passed through a fully-convolutional network to produce a 3D regular fixed grid $\bG^t$. \textbf{(b)} The $H \times W \times D$ grid showing cells responsible for recognizing hands and objects. \textbf{(c)} Each cell predicts the 3D hand pose and object bounding box coordinates in a 3D grid. \textbf{(d)} The output tensor from our network, in which the target values for hands and objects are stored. \textbf{(e)} Cells are associated with a vector that contains target values for hand and object pose, object and  action categories and an overall confidence value. Predictions with high confidence values are then passed through the \emph{interaction RNN} to propagate information in the temporal domain and model interactions in 3D between hands and objects.}
	\label{fig:approach}
\end{figure*}

\subsection{Overview}
\label{ssec:overview}

The general overview of our \emph{Hand+Object} model is given in Fig.~\ref{fig:approach}. Our model takes as input a sequence of color frames $\bI^t$ ($1 \leq t \leq N$) and predicts \emph{per-frame} 3D hand and object poses, object classes and action categories, along with \emph{per-sequence} interaction classes. Here, we define actions as \emph{verbs}, \eg ``pour'', and interactions as~\emph{(verb, noun)} pairs, \eg ``pour juice''. We represent hand and object poses with $N_c$ 3D control points. Details about control point definitions are provided in Sec.~\ref{sec:singleshot}. We denote the number of object classes by $N_o$, the number of actions by $N_{a}$ and the number of interactions by $N_{ia}$.

Our model first processes each frame, $\bI^t$, of a sequence with a fully convolutional network (Fig.~\ref{fig:approach}a) and divides the input image into a regular grid $\bG^t$ containing $H \times W \times D$ cells that span the 3D scene in front of the camera (Fig.~\ref{fig:approach}b). We keep the target values of our network for hands and objects in tensor $\bG^t$ (Fig.~\ref{fig:approach}c-d). Namely, the target values for a hand or an object at a specific cell location,  $i \in H \times W \times D$, are placed in the $i$-th cell of $\bG^t$ in the form of a multi-dimensional vector, $\bv_i$. To be able to jointly estimate the pose of a hand and an object potentially occluding each other, we allow each cell to store two separate sets of values, one for the hand, denoted by $\bv_i^h$, and one for the object, denoted by $\bv_i^o$ (Fig.~\ref{fig:approach}e).
Vector $\bv_i^h$ stores the control points for hand pose, $\by_{i}^h \in \real^{3 N_c}$, action probabilities, $\bp_{i}^{a} \in \real^{N_a}$, and an overall confidence value for the hand pose estimation, $c_{i}^h \in [0,1]$. Vector $\bv_i^o$ stores the control points for object pose, $\by_{i}^o \in \real^{3 N_c }$, object class probabilities, $\bp_{i}^o \in \real^{N_o}$, and an overall confidence value for the object pose estimation, $c_{i}^o \in [0,1]$.

We train our single pass network based on~\cite{Redmon17} to be able to predict these target values. At test time, predictions at cells with low confidence values, \ie where the hands or objects of interest are not present, are pruned. All these predictions are obtained with a single forward pass in the network. While very efficient, this step works on each frame independently, thus ignoring the temporal dimension. Therefore, we add a recurrent module to integrate information across frames and model the interaction between hands and objects (Fig.~\ref{fig:approach}a). This module takes as input hand and object predictions with high confidence values, and outputs a probability vector, $\bp^{ia} \in \real^{N_{ia}}$, over interaction classes. In the following sections, we describe each of these components in more detail.

\subsection{Joint 3D Hand-Object Pose Estimation}
\label{sec:singleshot}

In the context of rigid object pose estimation,~\cite{Rad17,Tekin18} regress the 2D locations of $8$ keypoints -- namely, the projections of the $8$ corners of the object's 3D bounding box. The object's 6D pose is then estimated using a PnP algorithm~\cite{Lepetit09}. Adopting a similar approach would not work in our case, since we aim at estimating also the articulated 3D pose of the hand. To tackle this problem and jointly estimate 3D articulated and rigid pose within the same architecture, we propose to use a common output representation for both hands and objects. To this end, we parametrize both hand and object poses jointly with 3D control points, corresponding to 21 skeleton joints for the hand pose and 3D locations of object keypoints, corresponding to the locations on the 3D object bounding box. For simplicity, we choose $N_c = 21$, and define 8 keypoints for the objects as proposed in~\cite{Rad17,Tekin18}, along with the 12 edge midpoints and the centroid of the 3D bounding box.

Adopting a coherent parameterization for hand and object simplifies the regression task. We subdivide the input image into a grid of $H \times W$ cells, and further discretize depth into $D$ cells. Note that discretization is defined in pixel space for the first two dimensions, and in metric space for depth. Therefore each cell has a size of $C_u \times C_v$ pixels, $\times C_z$ meters. Within each cell, we predict offsets $\Delta u$, $\Delta v$, $\Delta z$ for the locations corresponding to the control points with respect to the top-left corner of the cell that is closer to the camera, $(u, v, z)$. For the hand root joint and the object centroid, we constrain the offset to lie between $0$ and $1$, where a size of $1$ corresponds to the full extent of an individual cell within grid dimensions. For other control points, we do not constrain the network’s output as those points should be allowed to fall outside the cell. The predicted location of the control point $(\hat{w}_u, \hat{w}_v, \hat{w}_z)$ is then computed as:
\begin{align}
& \hat{w}_{u} = g(\Delta u) + u \\
& \hat{w}_{v} = g(\Delta v) + v \\ 
& \hat{w}_{z} = g(\Delta z) + z 
\label{eq:offsets}
\end{align}
\noindent where $g(\cdot)$ is chosen to be a 1D sigmoid function for the root joint and the object centroid, and the identity function for other control points. Here, $(u, v, z)$ are indices for the cell in grid dimensions. Given the camera intrinsics matrix $K$, and the prediction for the grid location, $(\hat{w}_u, \hat{w}_v, \hat{w}_z)$, the 3D location $\hat{\by}$ of the control point in the camera coordinate system is then computed as:
\begin{equation}
\hat{\by} = \hat{w}_z \cdot C_z \cdot K^{-1}[\hat{w}_u \cdot C_u, \hat{w}_v \cdot C_v, 1]^T.
\end{equation}
3D joint predictions already define the hand pose. Given the control point predictions on the 3D bounding box, 6D object pose could also be computed efficiently by aligning the prediction to the reference 3D bounding box with a rigid transformation. This dispenses with the need to solve for a 2D-to-3D correspondence problem using PnP as in~\cite{Rad17,Tekin18} and recovers the 6D pose via Procrustes transformation~\cite{Gower75}. Such a formulation also reduces depth ambiguities caused by projection from 3D to 2D. We show in Sec.~\ref{sec:results} that this results in an improved object pose estimation accuracy. 

In addition to hand and object control point locations, our network also predicts high confidence values for cells where the hand (or the object) is present, and low confidence where they are not present. Computing reliable confidence values is key for obtaining accurate predictions at test time.
We define the confidence of a prediction as a function of the distance of the prediction to the ground truth, inspired by~\cite{Tekin18}. Namely, given a predicted 2D location $(\hat{w}_u, \hat{w}_v)$ and its Euclidean distance $D_T(\hat{w}_u, \hat{w}_v)$ from the ground truth, measured in image space, the confidence value $c_{uv}(\hat{w}_u, \hat{w}_v)$ is computed as an exponential function with cut-off value $d_{th}$ and sharpness parameter $\alpha$: 
\begin{equation}
\label{eq:confidence}
c_{uv}(\hat{w}_u, \hat{w}_v) = e^{\alpha \left( 1 - \frac{D_T(\hat{w}_u, \hat{w}_v)}{d_{th}} \right)}
\end{equation}
\noindent if $D_T(\hat{w}_u, \hat{w}_v) < d_{th}$, and $c_{uv}(\hat{w}_u, \hat{w}_v)=0$ otherwise. However, in contrast to~\cite{Tekin18} that computes a confidence value for 2D prediction, we need to also consider the depth dimension. Therefore, for a given depth prediction, $\hat{w}_z$, we compute an additional confidence value, $c_{z}(\hat{w}_z)$, measuring the distance (in metric space) between prediction and ground truth as in Eq.~\ref{eq:confidence}. We then compute the final confidence value $c = 0.5 \cdot c_{uv}(\hat{w}_u, \hat{w}_v) + 0.5 \cdot c_{z}(\hat{w}_z)$.

\subsection{Object and Action Recognition}

In addition to the 3D control locations and the confidence value, our model also predicts the object and action classes (\ie \emph{nouns} and \emph{verbs}, as defined in Sec.~\ref{ssec:overview}). Intuitively, features learned to predict hand and object pose could also help recognize actions and object classes. In order to predict  actions, at each cell $i$ we store into vector ${\bf v}_i^h$, together with hand pose ${\bf y}_i^h$ and the corresponding confidence value $c^h_i$, the target action class probabilities ${\bf p}^{a}_i$. Similarly, to be able to predict object classes, we store into vector ${\bf v}_i^o$, together with object pose ${\bf y}_i^o$ and the corresponding confidence value $c^o_i$, the target object class probabilities ${\bf p}^{o}_i$.

In total, $\bv_i^h$ stores $3 \cdot N_c + N_{a} + 1$ values and $\bv_i^o$ stores $3 \cdot N_c + N_o + 1$ values. We train our network to be able to predict $\bv_i^h$ and $\bv_i^o$ for each cell $i$ of the 3D grid ${\bf G}^t$, for each $t$. Given \emph{verb} and \emph{noun} predictions, our model is able to recognize interactions from only a single image. Ultimately, our network learns to jointly predict 3D hand and object poses along with object, action and interaction classes -- with a single forward pass.

\vspace{-1mm}
\subsection{Temporal Reasoning and Interaction Modeling}
\label{sec:rnn}
While simple actions can be recognized by looking at single frames, more complex activities require to model longer-term dependencies across sequential frames. To reason along the temporal dimension, we add a RNN module to our architecture. We choose to use a Long Short-Term Memory (LSTM)~\cite{Hochreiter1997}, given its popularity in the action recognition literature~\cite{GarciaHernando18}. We also experimented with different modules (\eg Gated Recurrent Units~\cite{cho14}), without observing substantial differences in our results.

A straightforward approach would be to consider the highest confidence predictions for hand and object poses at each frame, and give them as input to the RNN module, as in~\cite{GarciaHernando18}. The RNN output is then processed by a softmax layer to predict the activity class. However, we can improve upon this baseline by explicitly reasoning about \emph{interactions} between hands and objects.  For example, in the context of visual reasoning,~\cite{Baradel18} proposes to model relational dependencies between objects in the scene and demonstrates that such object-level reasoning improves accuracy in targeted visual recognition tasks. In a similar manner, we aim to explicitly model interactions -- in this case, however, of hands and objects, and directly in 3D. Co-training of 3D hand and object pose estimation networks already implicitly accounts for interactions in a data-driven manner. We further propose to model hand-object interactions at the structured output level with an \emph{interaction RNN}. To do so, instead of directly feeding the hand and object poses as input to the temporal module, inspired by~\cite{Baradel18,Santoro17}, we model dependencies between hands and objects with a composite learned function and only then give the resulting mapping as input to RNN:

\vspace{-3mm}
\begin{equation}
f_{\phi}(g_{\theta}(\hat{{\bf y}}^h,\hat{{\bf y}}^o))
\end{equation}
\vspace{-3mm}

\noindent where $f_{\phi}$ is an LSTM and $g_{\theta}$ is an MLP, parameterized by $\phi$ and $\theta$, respectively.
Ultimately this mapping learns the explicit dependencies between hand and object poses and models interactions. We analyze and quantitatively evaluate the benefits of this approach in Sec.~\ref{sec:results}.

\subsection{Training}
\label{sec:training}
\vspace{-1mm}
The final layer of our single-shot network produces, for each cell $i$, hand and object pose predictions, $\hat{\by}^h_{i}$ and $\hat{\by}^o_{i}$,
with their overall confidence values, $\hat{c}^h_i$ and $\hat{c}^o_i$, as well as probabilities for actions, $\hat{\bp}^{a}_i$, and object classes, $\hat{\bp}^{o}_i$.
For each frame $t$, the loss function to train our network is defined as follows:
\begin{align}
\mathcal{L} \;\; = & \;\; \lambda_{pose} \sum_{i \in {\bG}^t} (|| \hat{\by}^h_{i} - \by^h_{i} ||_2^2 + || \hat{\by}_{i}^o - \by^o_{i} ||_2^2) + \\
& \;\; \lambda_{conf} \sum_{i \in {\bG}^t} ( (\hat{c}^h_i - c^h_i)^2 + (\hat{c}^o_i - c^o_i)^2 ) - \\
& \;\; \lambda_{actcls} \sum_{i \in {\bG}^t} \hat{\bp}_{i}^{a} \log \bp_{i}^{a} - \\
& \;\; \lambda_{objcls} \sum_{i \in {\bG}^t} \hat{\bp}_{i}^{o} \log \bp_{i}^{o}.
\end{align}
\vspace{-3mm}

Here, the regularization parameters for the pose and classification losses, $\lambda_{pose}$, $\lambda_{actcls}$ and $\lambda_{objcls}$ are simply set to $1$. As suggested by~\cite{Redmon17}, for cells that contain a hand or an object, we set $\lambda_{conf}$ to $5$ and for cells that do not contain any of them, we set it to $0.1$ to increase model stability. We feed the highest confidence predictions of our network to the recurrent module and define an additional loss based on cross entropy for the action recognition over the entire sequence. In principle, it is straightforward to merge and train the single-image and temporal models jointly with a softargmax operation~\cite{Chapelle09,Yi16}.
However, as backpropagation requires to keep all the activations in memory, which is not possible for a batch of image sequences, we found it effective to train our complete model in two stages. We first train on single frames to jointly predict 3D hand and object poses, object classes and action categories. We then keep the weights of the initial model fixed and train our recurrent network to propagate information in the temporal domain and model interactions. The complete model takes as input a sequence of images and outputs per-frame 3D hand-object pose predictions, object  and action classes along with the estimates of interactions for the entire sequence.

\vspace{-2mm}
\section{Evaluation}
\label{sec:results}
\vspace{-2mm}
In this section, we first describe the datasets and the corresponding evaluation protocols. We then compare our \emph{Hand+Object} approach against the state-of-the-art methods and provide a detailed analysis of our general framework.
\vspace{-1mm}
\subsection{Datasets}
\label{ssec:datasets}

We evaluate our framework for recognizing 3D hand-object poses and interactions on the recently introduced First-Person Hand Action (FPHA) dataset~\cite{GarciaHernando18}. It is the only publicly available dataset for 3D hand-object interaction recognition that contains labels for 3D hand pose, 6D object pose and action categories. FPHA is a large and diverse dataset including $1175$  videos belonging to $45$ different activity categories performed by 6 actors. A total of 105,459 frames are annotated with accurate hand poses and action categories.  The subjects carry out complex motions corresponding to daily human activities. A subset of the dataset contains annotations for objects' 6-dimensional poses along with corresponding mesh models for 4 objects involving 10 different action categories. We denote this subset of the dataset as FPHA-HO.

As there are no other egocentric datasets containing labels for both 3D hand pose and 6D object pose, we further annotate 6D object poses on a part of the EgoDexter hand pose estimation dataset~\cite{Mueller17}, on which we validate our joint hand-object pose estimation framework. To this end, we annotate the \emph{Desk} sequence between frames $350$ and $500$ which features a cuboid object.
We use training data from the SynthHands 3D hand pose estimation dataset~\cite{Mueller17}. We augment this dataset by randomly superimposing on the image synthetic cuboid objects that we generate and object segmentation masks from the third-person view dataset of~\cite{Sridhar16} which features the same object. To gain robustness against background changes, we further replace the backgrounds using random images from~\cite{Everingham10}. 

\subsection{Evaluation Metrics}

We evaluate our unified framework on a diverse set of tasks: egocentric activity recognition, 3D hand pose estimation and 6D object pose estimation, and use standard metrics and official train/test splits to evaluate our performance in comparison to the state of the art. We use the percentage of correct video classifications, percentage of correct keypoint estimates (3D PCK) and percentage of correct poses to measure accuracy on activity recognition, 3D hand pose estimation and 6D object pose estimation, respectively. When using the 3D PCK metric for hand pose estimation, we consider a pose estimate to be correct when the mean distance between the predicted and ground-truth joint positions is less than a certain threshold. When using the percentage of correct poses to evaluate 6D object pose estimation accuracy, we take a pose estimate to be correct if the 2D projection error or the average 3D distance of model vertices is less than a certain threshold (the latter being also referred to as the ADD metric).

\subsection{Implementation Details}

We initialize the parameters of our single-image network based on~\cite{Redmon17} with  weights pretrained on ImageNet. The input to our model is a $416 \times 416$ image. The output grid, $\bG^t$, has the following dimensions: $H=13$, $W=13$ and $D=5$. We set the grid size in image dimensions to $C_u = C_v = 32$ pixels as in~\cite{Redmon17} and in depth dimension to $C_z = 15$ cm. Further details about the architecture can be found in the supplemental material. We set the sharpness of the confidence function, $\alpha$, to $2$, and the distance threshold to $75$ pixels for the spatial dimension and $75$ mm for the depth dimension. We use a 2-layer LSTM with a hidden layer size of $512$. The nonlinearity, $g_{\theta}$, is implemented as an MLP with 1 hidden layer with ReLU activation consisting of $512$ units. We use stochastic gradient descent for optimization. We start with a learning rate of $0.0001$ and divide the learning rate by $10$ at the $80^{th}$ and $160^{th}$ epoch. All models are trained with a batch size of $16$ for $200$ epochs. We use extensive data augmentation to prevent overfitting. We randomly change the hue, saturation and exposure of the image by up to a factor of $50 \%$. We also randomly translate the image by up to a factor of $10\%$. 
Our implementation is based on PyTorch.

\subsection{Experimental Results}

We first report activity recognition accuracy on the FPHA-HO dataset and compare our results to the state-of-the-art results  of~\cite{GarciaHernando18} in Table~\ref{tab:actionrecognitionsubset}. We further use the following baselines and versions of our approach in the evaluation:
\vspace{-5mm}
\begin{itemize}
\item \textsc{Single-Image} - Our single pass network that predicts the  action (\ie \emph{verb}), and object class (\ie \emph{noun}). The individual predictions for  action (${\bf p}^{a}_i$, \eg open) and object (${\bf p}^{o}_i$, \eg bottle) class are combined to predict the interaction type (\eg open bottle), \ie \emph{(verb, noun)} pair. This version of our model does not use a temporal model.
\vspace{-2mm}
\item \textsc{Hand Pose} - A temporal model that uses the hand pose predictions of our approach as input to the RNN to recognize activities.
\vspace{-2mm}
\item \textsc{Object Pose} - similar to the previous baseline, but trained to predict activities based on the predicted keypoints on the 3D object bounding box.
\vspace{-2mm} 
\item \textsc{Hand + Object Pose} - A version of our model that combines 3D hand-object pose predictions to feed them as input to the RNN.
\vspace{-2mm}
\item \textsc{Hand Pose + Object Pose + Interact} - A complete version of our model with temporal reasoning and interaction modeling.
\end{itemize}

\vspace{-4.45mm}

\begin{table}[b]
	\centering
	\tabcolsep=0.1cm
	\scalebox{0.81}{
		\begin{tabular}[b]{l|lclc}
			\hline
			Method 				 &Model	 	              					 & Action Accuracy (\%)  \\
			\hline
			\multirow{3}{*}{\cite{GarciaHernando18}}&Ground-truth Hand Pose  & 87.45  \\
			&Ground-truth Object Pose     									 & 74.45  \\
			&Ground-truth Hand + Object Pose 	  							 & 91.97  \\
			\hline
			\multirow{5}{*}{OURS}&\textsc{Single-Image}					     & 85.56   \\
			&\textsc{Hand Pose} 	  					  					 & 89.47   \\
			&\textsc{Object Pose}											 & 85.71    \\
			&\textsc{Hand + Object Pose}									 & 94.73     \\
			&\textsc{Hand + Object Pose + Interact}               			 & \textbf{96.99}     \\
			\hline
		\end{tabular}
	}
	\caption{Action recognition results on FPHA-HO. We evaluate the impact of hand and object poses for action recognition. We demonstrate that hand-object predictions of our unified network yield more accurate results than~\cite{GarciaHernando18} which relies on ground-truth pose as input. Furthermore, explicitly modeling interactions of hand-object poses results in a clear improvement in accuracy.}
	\label{tab:actionrecognitionsubset}
\end{table}

\begin{figure}[t]
	\centering
	\scalebox{0.8}{
		\begin{tabular}{cc}
			\hspace{-3mm}\includegraphics[width=0.55\columnwidth,height=1.4in]{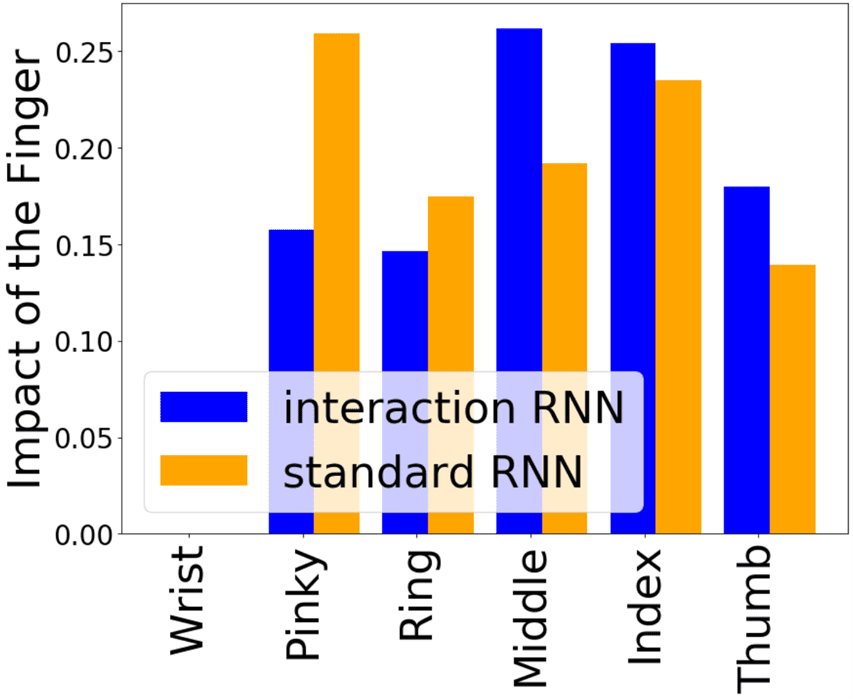}
			& \includegraphics[width=0.55\columnwidth,height=1.4in]{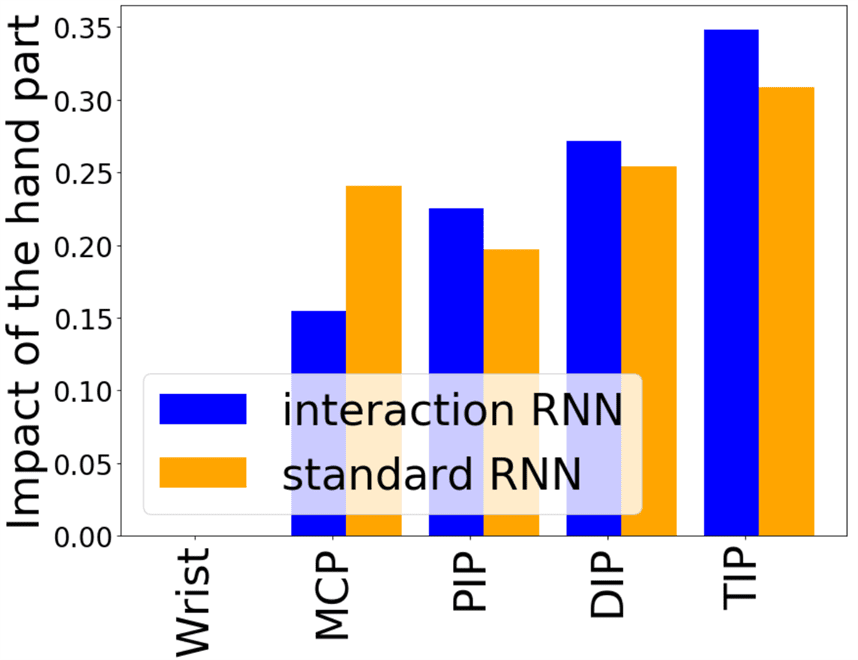} \\ \vspace{-7mm}
	\end{tabular}}
	\caption{The impact of each finger (left) and each hand part (right) in recognizing interactions. The impact is measured by the normalized magnitude of neural network weights tied to the corresponding hand joint positions and compared for a standard RNN and an interaction RNN. MCP, PIP and DIP denote the 3 consecutive joints located inbetween wrist and fingertip (TIP) on each finger, in their respective order.}\vspace{-2mm}
	\label{fig:interactionanalysis}
\end{figure}

\begin{figure*}[t]
	\centering
	\scalebox{0.41}{
		\begin{tabular}{ccccc}
			\hspace{-0.1cm} \includegraphics[width=0.34\linewidth,height=1.7in]{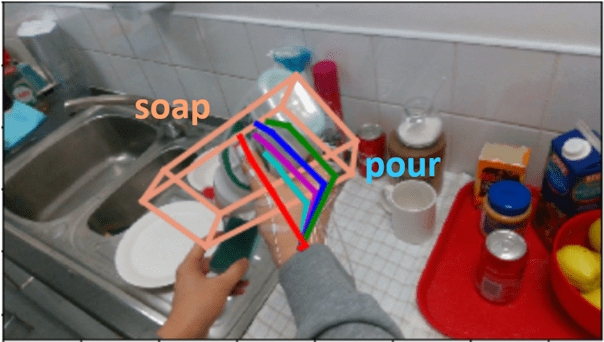} \hspace{-3mm}
			&\includegraphics[width=0.34\linewidth,height=1.7in]{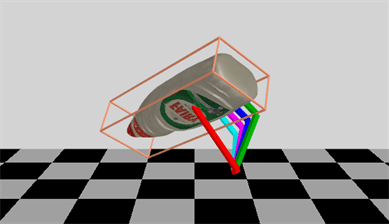} \hspace{0.5cm}
			&\includegraphics[width=0.34\linewidth,height=1.7in]{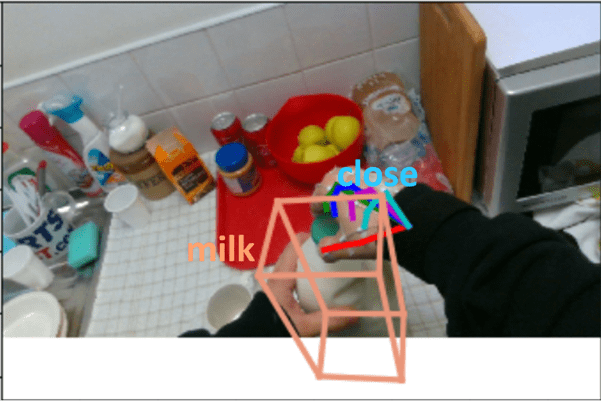} \hspace{-3mm}
			&\includegraphics[width=0.34\linewidth,height=1.7in]{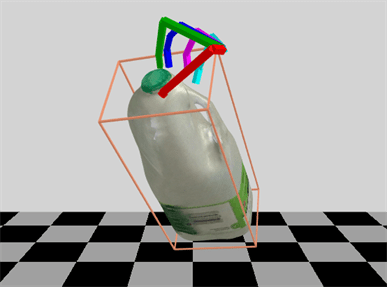}\hspace{0.5cm}
			&\includegraphics[width=0.72\linewidth,height=1.7in]{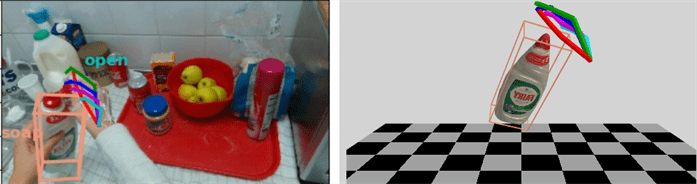} \\
			\hspace{-0.1cm}\includegraphics[width=0.34\linewidth,height=1.7in]{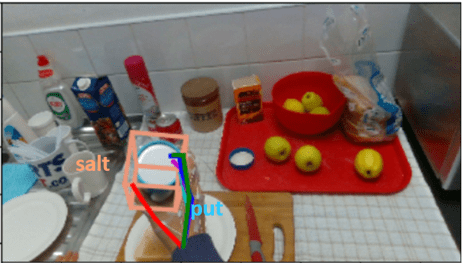}\hspace{-3mm}
			&\includegraphics[width=0.34\linewidth,height=1.7in]{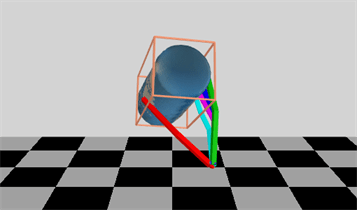}\hspace{0.5cm}
			&\includegraphics[width=0.34\linewidth,height=1.7in]{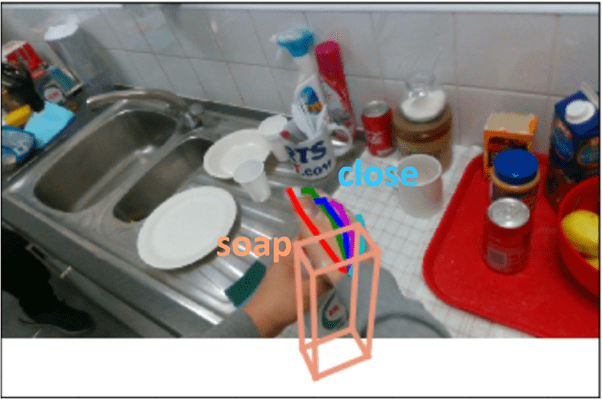}\hspace{-3mm}
			&\includegraphics[width=0.34\linewidth,height=1.7in]{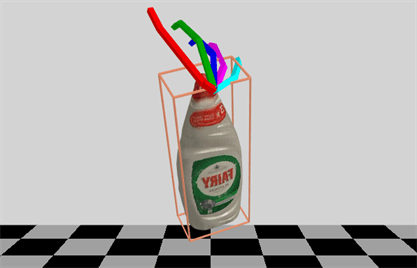} \hspace{0.5cm}
			&\includegraphics[width=0.72\linewidth,height=1.7in]{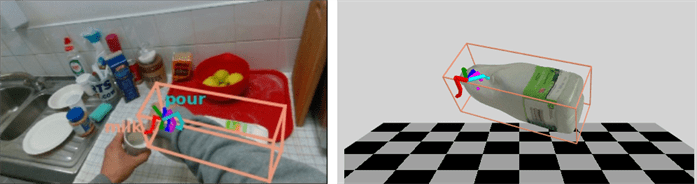} \\
			\hspace{-0.1cm}\includegraphics[width=0.34\linewidth,height=1.7in]{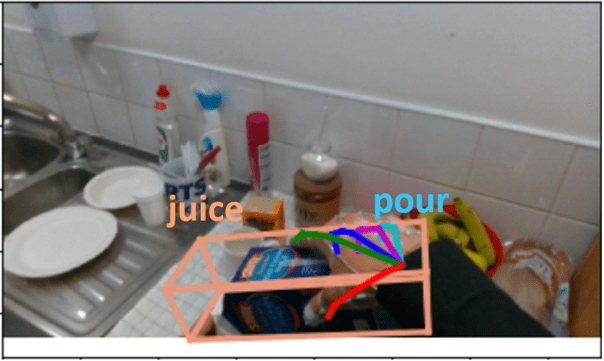}\hspace{-3mm}
			&\includegraphics[width=0.34\linewidth,height=1.7in]{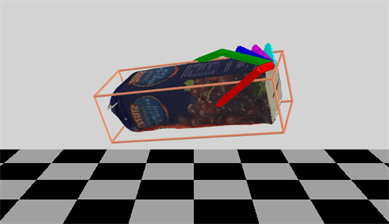}\hspace{0.5cm}
			&\includegraphics[width=0.34\linewidth,height=1.7in]{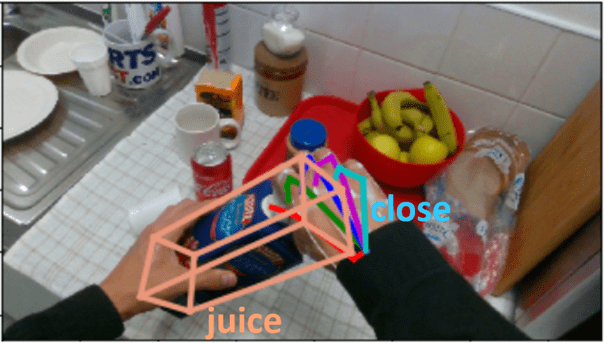}\hspace{-3mm}
			&\includegraphics[width=0.34\linewidth,height=1.7in]{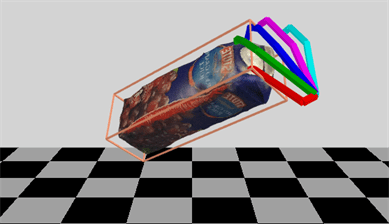} \hspace{0.5cm}
			&\includegraphics[width=0.72\linewidth,height=1.7in]{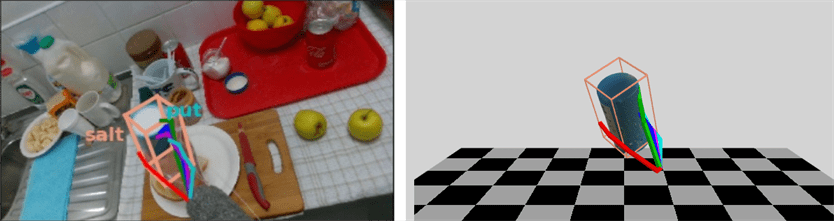}\\
			\hspace{-0.1cm}\includegraphics[width=0.34\linewidth,height=1.7in]{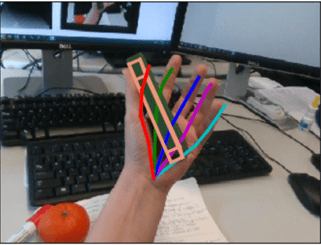}\hspace{-3mm}
			&\includegraphics[width=0.34\linewidth,height=1.7in]{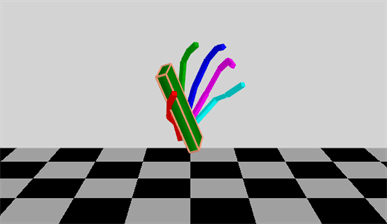}\hspace{0.5cm}
			&\includegraphics[width=0.34\linewidth,height=1.7in]{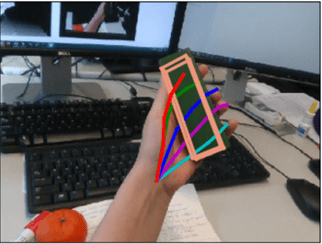}\hspace{-3mm}
			&\includegraphics[width=0.34\linewidth,height=1.7in]{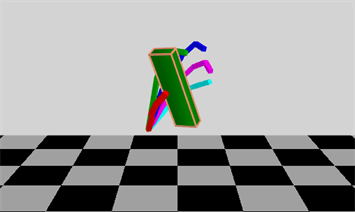}\hspace{0.5cm}
			&\includegraphics[width=0.72\linewidth,height=1.7in]{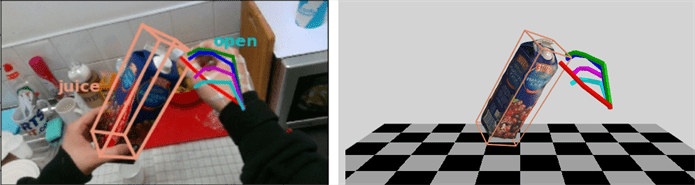}  \\	
		\end{tabular}
	} \vspace{0mm}
	\caption{Qualitative  results on the FPHA and EgoDexter dataset. We visualize the 3D hand pose estimates, 3D object bounding boxes which are transformed with the learned 6D object poses, and activity labels. The proposed approach can handle motion blur, self-occlusions, occlusions by viewpoints, clutter and complex articulations. We provide further qualitative results in our supplemental material.}
	\label{fig:results}
\end{figure*}

\paragraph{Recognizing Interactions.} Our model achieves state-of-the-art performance for recognizing egocentric activities and 3D hand-object interactions even without ground-truth 3D hand and object poses. Our \textsc{Single-Image} baseline already achieves close results to the state of the art. We demonstrate that temporal reasoning on 3D hand and object poses individually improves activity recognition accuracy. The combination of hand and object poses further boosts the overall scores. With interaction modeling, performance improvements are even more noticeable. To analyze the importance of interaction modeling further, in Fig.~\ref{fig:interactionanalysis}, we quantify the importance of each input to the RNN by measuring the magnitude of the network weights tied to the inputs, both for a standard RNN and our interaction RNN. We demonstrate that, in contrast to a standard RNN, our temporal model attributes more importance to the index fingers and fingertips that interact more commonly with objects and  learns which joints are relevant in interaction. In Fig.~\ref{fig:results}, we provide some visual results, 3D hand and object poses along with action and object labels that are produced by our \textsc{Hand + Object + Interact} model.

 We further evaluate the performance of our approach for the task of egocentric activity and interaction recognition on the full FPHA dataset. On the full dataset, object poses are not available for all the action categories, therefore, we train our models only using 3D hand poses. Here, however, to increase the descriptive power, we further leverage the object class and  action category predictions produced by our single pass network  in our temporal model. To this end we augment the 3D hand pose input (HP) with the output object class (OC) and  action category (AC) probabilities. In Table~\ref{tab:actionrecogntionfull}, we demonstrate that these additional features yield improved action recognition accuracy. Overall, our method consistently outperforms the baselines by a large margin, including the ones that rely on depth data.

\begin{figure}[t]
	\centering
	\scalebox{0.52}{
		\begin{tabular}{c}
			\includegraphics[width=\columnwidth]{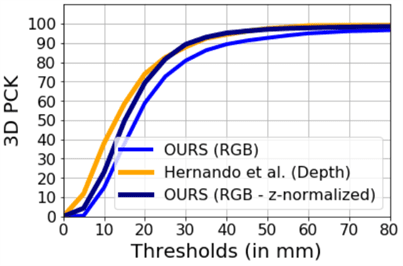}
	\end{tabular}}\vspace{-2mm}
	\caption{Comparison of the hand pose estimation results of our approach with those of Garcia-Hernando et al.~\cite{GarciaHernando18} using different thresholds for the 3D pose error.}\vspace{-1mm}
	\label{fig:hpecomparison}
\end{figure}

\begin{table}[tbph]
	\centering
	\tabcolsep=0.1cm
	\scalebox{0.76}{
		\begin{tabular}[b]{lcc}
			\hline
			Model 				            		  &Input modality & Accuracy  \\
			\hline
			Two-stream-color~\cite{Feichtenhofer16}	  &Color		    &61.56 \\
			Two-stream-flow~\cite{Feichtenhofer16}    &Color	     	&69.91 \\
			Two-stream-all~\cite{Feichtenhofer16}	  &Color			&75.30 \\
			Joule-color~\cite{Hu15}		  			  &Color			&66.78 \\
			HON4D~\cite{Oreifej13}				  	  &Depth			&70.61 \\
			Novel View~\cite{Rahmani16}			  	  &Depth			&69.21 \\
			Joule-depth~\cite{Hu15}		  			  &Depth			&60.17 \\ 
			\cite{GarciaHernando18} + Gram Matrix     &Depth			&32.22 \\
			\cite{GarciaHernando18} + Lie Group       &Depth			&69.22 \\
			\cite{GarciaHernando18} + LSTM			  &Depth            &72.06 \\
			\hline
			OURS - HP  								  &Color            &62.54\\ 
			OURS - HP + AC 							  &Color            &74.20\\ 
			OURS - HP + AC + OC 					  &Color            &\textbf{82.43}\\ 
			\hline
		\end{tabular}
	}
	\caption{Action recognition results on the full FPHA dataset~\cite{GarciaHernando18}. Our method significantly improves upon the baselines, including the ones that rely on depth data. We further provide action-specific recognition accuracies in the supplemental material.}
	\vspace{-2mm}
	\label{tab:actionrecogntionfull}
\end{table}

\vspace{-4mm}
\paragraph{3D Hand Pose Prediction.} We further compare the accuracy of our 3D hand pose predictions to the state-of-the-art results on FPHA in Fig.~\ref{fig:hpecomparison}. Even though we only use \emph{color} images, in contrast to~\cite{GarciaHernando18} that uses \emph{depth} input, we achieve competitive 3D hand pose estimation accuracy. Furthermore, we do not assume knowledge of the hand bounding box as in~\cite{GarciaHernando18}; our model takes as input a single full color image. Note also that the method of~\cite{GarciaHernando18} is specifically trained for 3D hand pose estimation, whereas this is a subtask of our method which simultaneously tackles multiple tasks within a unified architecture. 

\begin{figure}[t]
	\centering
	\scalebox{0.47}{
		\begin{tabular}{cc}
			\hspace{-3mm}\includegraphics[width=\columnwidth]{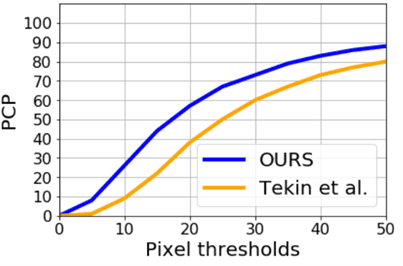}
			& \includegraphics[width=\columnwidth]{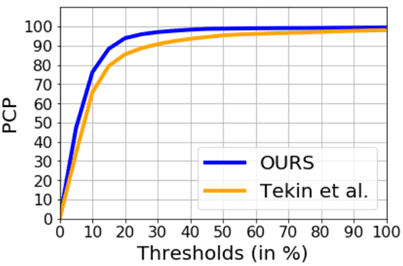} \\
	\end{tabular}}\vspace{-2mm}
	\caption{Comparison of the object pose estimation results of our approach with those of Tekin et al.~\cite{Tekin18} using different thresholds for the 2D projection (left) and ADD metric (right).} \vspace{-2mm}
	\label{fig:opecomparison}
\end{figure}

\vspace{-4mm}
\paragraph{6D Object Pose Prediction.}\hspace{-3mm} To evaluate our object pose accuracy, we compare our approach to the state-of-the-art results of~\cite{Tekin18} in Fig.~\ref{fig:opecomparison}. To this end, we run their approach on FPHA with their publicly available code. We demonstrate that explicitly reasoning about 6D object pose in 3D, in contrast to~\cite{Tekin18} that relies on solving 2D-to-3D correspondences, yields improved pose estimation accuracy. We conjecture that posing the 6D pose estimation problem in 2D is prone to depth ambiguities and our approach brings in robustness against it by directly reasoning in 3D.

\vspace{-4mm}
\paragraph{Unified Framework.} We analyze the influence of simultaneously training hand and object pose estimation tasks within the same single pass network on individual pose estimation accuracies. We compare the results of our \emph{Hand + Object} network to those of the networks trained only for hand pose estimation and only for object pose estimation in Table~\ref{tab:unifiedvsindividual}. While we obtain similar accuracies for hand pose estimation with co-training and individual training, object pose estimation accuracy for the unified network is significantly better than that of the individual pose estimation network by a large margin of $9.65\%$. This indicates that having a joint representation which is shared across multiple tasks leads to an improvement in combined pose estimation accuracy. The results suggest that 3D hand pose highly constrains the 6D object pose, while the effect of rigid object pose on the articulated hand pose is not as pronounced. We further show that the prediction of the interaction class improves the hand and object pose estimation accuracy. This further validates that our unified framework allows us to achieve better overall performance in targeted tasks.

\vspace{-1mm}

\begin{table}[tbph]
	\centering
	\tabcolsep=0.1cm
	\scalebox{0.76}{
		\begin{tabular}[b]{lcc}
			\hline
			Network 									& HP error	 	       & OP error  \\
			\hline
			\textsc{Hand Only} 									&16.15   				     	& -     \\
			\textsc{Object only}									&-	  							& 28.27  \\
			\textsc{Hand + Object}								&16.87 	  					    & 25.54   \\
			\textsc{Hand + Object + Interact}					&\textbf{15.81} 	  			& \textbf{24.89}   \\
			\hline
		\end{tabular}
	}
	\caption{Comparison of the pose estimation results of our unified network to those of the networks trained only for hand and object pose estimation. Error metric is the mean 3D distance in mm.} \vspace{-3mm}
	\label{tab:unifiedvsindividual}
\end{table}

\vspace{-6mm}
\paragraph{Generalization.}

To demonstrate the generalization power of our joint hand-object pose estimation framework, we annotate a part of the EgoDexter hand pose estimation dataset~\cite{Mueller17} with 6D object poses, as explained in Sec.~\ref{ssec:datasets}, and report quantitative results in Table~\ref{tab:egodexter}. We demonstrate that even when trained on synthetic data, our approach generalizes well to unconstrained environments and results in reliable and accurate joint 3D hand-object pose estimates. We provide visual results on this dataset in Fig.~\ref{fig:results}.

\begin{table}[t]
	\centering
	\tabcolsep=0.1cm
	\scalebox{0.87}{
		\begin{tabular}[b]{lc}
			\hline
			Part									& Error (in cm)	 	  \\
			\hline
			Fingertips  							&4.84     \\
			Object coordinates						&2.37	  	  \\
			\hline
		\end{tabular}
	}
	\caption{Results on the augmented EgoDexter dataset. Even when trained on synthetic data, our approach yields accurate poses.}
	\vspace{-2mm}
	\label{tab:egodexter}
\end{table}

\vspace{-4mm}
\paragraph{Runtime.}

Our single pass network that produces per-frame predictions simultaneously for 3D hand poses, 6D object poses, object classes and action categories runs at real-time speed of $25$ fps on an NVIDIA Tesla M40. Without action and object recognition, when estimating only the poses of hands and objects, our method runs at a greater speed of $33$ fps. Given  hand and object poses, the interaction RNN module further processes a sequence with virtually no time overhead, at an average of $0.003$ seconds. 

\vspace{-3mm}

\section{Conclusion}
\label{sec:conclusion}
\vspace{-1mm}

In this paper, we propose the first method to jointly recognize 3D hand and object poses from  monocular color image sequences. Our unified \emph{Hand+Object} model simultaneously predicts per-frame 3D hand poses, 6D object poses, object classes and action categories, while being able to run at real-time speeds. Our framework jointly solves 3D articulated and rigid pose estimation problems within the same single-pass architecture and models the interactions between hands and objects in 3D to recognize actions from first-person views. Future work will apply the proposed framework to explicitly capture interactions between two hands and with other people in the scene.

{\small
\bibliographystyle{ieee}
\bibliography{ref}
}

\clearpage
\title{Supplemental Material: \\ H+O: Unified Egocentric Recognition of 3D Hand-Object Poses and Interactions}

\pagenumbering{gobble}


\author{Bugra Tekin$^1$ \quad\quad\quad\quad Federica Bogo$^1$ \quad\quad\quad\quad Marc Pollefeys$^{1,2}$ \\
	$^1$ Microsoft \quad\quad\quad\quad $^2$ ETH Z\"{u}rich
}

\maketitle


In the supplemental material, we provide details on the network architecture and how the training images were prepared. We also present additional qualitative and quantitative results on the datasets we evaluate our approach on~\cite{GarciaHernando18,Mueller17}.

\vspace{-4mm}
\paragraph{Network Architecture.}

Our unified network that jointly predicts per-frame 3D hand poses, 6D object poses, object classes and activity categories is a single pass network that does not rely on external detection algorithms nor region proposals~\cite{Redmon17}. To facilitate reproducibility, we provide the full details of our network architecture in Table~\ref{tab:networkarchitecture}. 

\vspace{-4mm}


\begin{table}[b]
	\begin{center}
		\scalebox{0.55}{
			\begin{tabular}{|c|c|c|c|c|c|}
				\hline
				Layer & Type & Filters & Size/Stride & Input & Output \\
				\hline
				0     &conv   &  32  &3 $\times$ 3 / 1   &416 $\times$ 416 $\times$    3     & 416 $\times$ 416 $\times$   32\\
				1     &max    &      &2 $\times$ 2 / 2   &416 $\times$ 416 $\times$   32     & 208 $\times$ 208 $\times$   32\\
				2     &conv   &  64  &3 $\times$ 3 / 1   &208 $\times$ 208 $\times$   32     & 208 $\times$ 208 $\times$   64\\
				3     &max    &      &2 $\times$ 2 / 2   &208 $\times$ 208 $\times$   64     & 104 $\times$ 104 $\times$   64\\
				4     &conv   & 128  &3 $\times$ 3 / 1   &104 $\times$ 104 $\times$   64     & 104 $\times$ 104 $\times$  128\\
				5     &conv   &  64  &1 $\times$ 1 / 1   &104 $\times$ 104 $\times$  128     & 104 $\times$ 104 $\times$   64\\
				6     &conv   & 128  &3 $\times$ 3 / 1   &104 $\times$ 104 $\times$   64     & 104 $\times$ 104 $\times$  128\\
				7     &max    &      &2 $\times$ 2 / 2   &104 $\times$ 104 $\times$  128     &  52 $\times$  52 $\times$  128\\
				8     &conv   & 256  &3 $\times$ 3 / 1   & 52 $\times$  52 $\times$  128     &  52 $\times$  52 $\times$  256\\
				9     &conv   & 128  &1 $\times$ 1 / 1   & 52 $\times$  52 $\times$  256     &  52 $\times$  52 $\times$  128\\
				10    &conv   & 256  &3 $\times$ 3 / 1   & 52 $\times$  52 $\times$  128     &  52 $\times$  52 $\times$  256\\
				11    &max    &      &2 $\times$ 2 / 2   & 52 $\times$  52 $\times$  256     &  26 $\times$  26 $\times$  256\\
				12    &conv   & 512  &3 $\times$ 3 / 1   & 26 $\times$  26 $\times$  256     &  26 $\times$  26 $\times$  512\\
				13    &conv   & 256  &1 $\times$ 1 / 1   & 26 $\times$  26 $\times$  512     &  26 $\times$  26 $\times$  256\\
				14    &conv   & 512  &3 $\times$ 3 / 1   & 26 $\times$  26 $\times$  256     &  26 $\times$  26 $\times$  512\\
				15    &conv   & 256  &1 $\times$ 1 / 1   & 26 $\times$  26 $\times$  512     &  26 $\times$  26 $\times$  256\\
				16    &conv   & 512  &3 $\times$ 3 / 1   & 26 $\times$  26 $\times$  256     &  26 $\times$  26 $\times$  512\\
				17    &max    &      &2 $\times$ 2 / 2   & 26 $\times$  26 $\times$  512     &  13 $\times$  13 $\times$  512\\
				18    &conv   &1024  &3 $\times$ 3 / 1   & 13 $\times$  13 $\times$  512     &  13 $\times$  13 $\times$ 1024\\
				19    &conv   & 512  &1 $\times$ 1 / 1   & 13 $\times$  13 $\times$ 1024     &  13 $\times$  13 $\times$  512\\
				20    &conv   &1024  &3 $\times$ 3 / 1   & 13 $\times$  13 $\times$  512     &  13 $\times$  13 $\times$ 1024\\
				21    &conv   & 512  &1 $\times$ 1 / 1   & 13 $\times$  13 $\times$ 1024     &  13 $\times$  13 $\times$  512\\
				22    &conv   &1024  &3 $\times$ 3 / 1   & 13 $\times$  13 $\times$  512     &  13 $\times$  13 $\times$ 1024\\
				23    &conv   &1024  &3 $\times$ 3 / 1   & 13 $\times$  13 $\times$ 1024     &  13 $\times$  13 $\times$ 1024\\
				24    &conv   &1024  &3 $\times$ 3 / 1   & 13 $\times$  13 $\times$ 1024     &  13 $\times$  13 $\times$ 1024\\
				25    &route  &16	 &			         &					                 &							     \\
				26    &conv   &  64  &1 $\times$ 1 / 1   & 26 $\times$  26 $\times$ 512      & 26 $\times$  26 $\times$  64	 \\
				27    &reorg  &      & / 2               & 26 $\times$  26 $\times$  64      & 13 $\times$  13 $\times$ 256	 \\
				28    &route  &27 24 &			         &					                 &							     \\
				29    &conv   &1024  &3 $\times$ 3 / 1   & 13 $\times$  13 $\times$ 1280     & 13 $\times$  13 $\times$ 1024 \\
				30    &conv   & 720  &1 $\times$ 1 / 1   & 13 $\times$  13 $\times$ 1024     & 13 $\times$  13 $\times$ 10 $\cdot$ (3$\times$ $N_c$+1+$N_a$+$N_o$)	 \\
				31    &prediction &  &                   &                                   & 13 $\times$  13 $\times$ 5 $\times$ 2 $\times$ (3$\times N_c$+1+$N_a$+$N_o$) \\
				\hline
		\end{tabular}}
	\end{center}
	\caption{Network architecture}
	\label{tab:networkarchitecture}
\end{table}

\begin{figure}[tbph]
	\centering
	\scalebox{0.6}{
		\begin{tabular}{cc}
			\includegraphics[height=3.2cm,width=3.2cm]{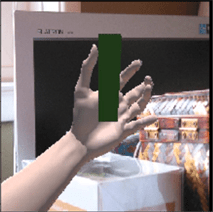}
			&\includegraphics[height=3.2cm,width=3.2cm]{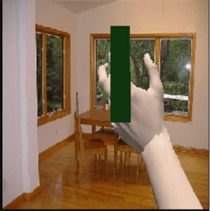}\\
			\includegraphics[height=3.2cm,width=3.2cm]{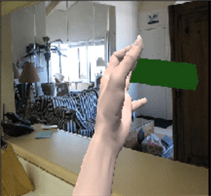}
			& \includegraphics[height=3.2cm,width=3.2cm]{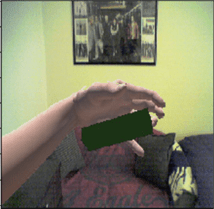}\\
		\end{tabular}
	}
	\caption{We extract the foreground objects in our training images and composite them over random images from PASCAL VOC~\cite{Everingham10}. We also augment the training set by superimposing segmentation masks of objects of interest for which the 6D poses are known to be able to simultaneously predict the hand and object pose and gain robustness against object occlusions.}\vspace{-7mm}
	\label{fig:trainingimages}
\end{figure}

\paragraph{Training Images.} As discussed in the main paper, while training on the synthetic hand data~\cite{Mueller17}, we replace the chroma-keyed background with random images from the PASCAL VOC dataset~\cite{Everingham10}. This operation of using random backgrounds brings in robustness against different backgrounds and is essential to achieve proper generalization.  In addition, we superimpose synthetic objects (cuboids) with known 6D poses on the training images. This allows us to train a network for both hand and object pose estimation and gain robustness against object occlusions. Examples of such images, which are given as input to the network at training time are shown in Fig.~\ref{fig:trainingimages}.

\begin{figure*}[t]
	\centering
	\scalebox{0.94}{
		\begin{tabular}{cc}
			\vspace{1cm}			
			\includegraphics[width=0.5\linewidth]{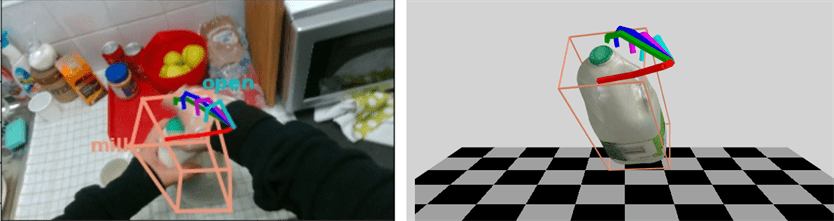}
			&\includegraphics[width=0.5\linewidth]{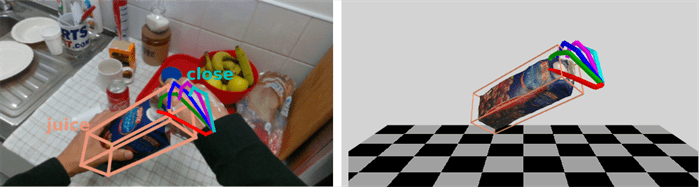} \\ \vspace{7mm}
			\includegraphics[width=0.5\linewidth]{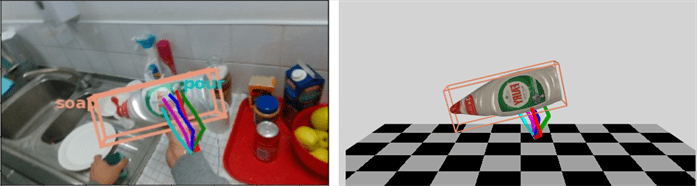}
			&\includegraphics[width=0.5\linewidth]{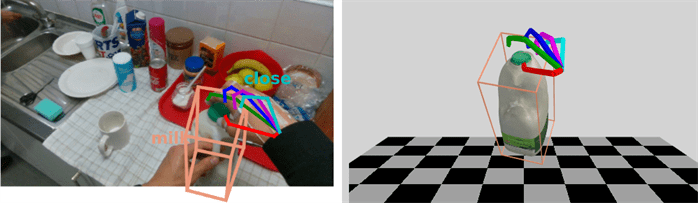} \\ \vspace{7mm} 
			\includegraphics[width=0.5\linewidth]{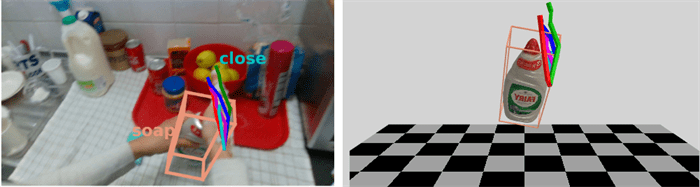}
			&\includegraphics[width=0.5\linewidth]{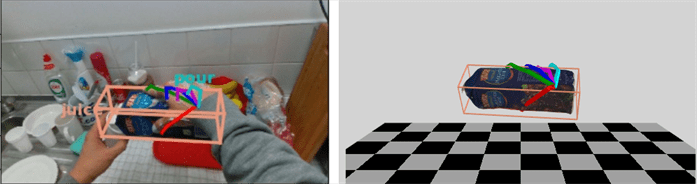} \\ \vspace{7mm}
			\includegraphics[width=0.5\linewidth]{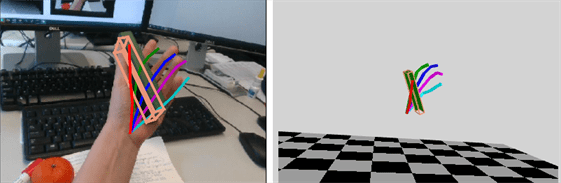}
			&\includegraphics[width=0.5\linewidth]{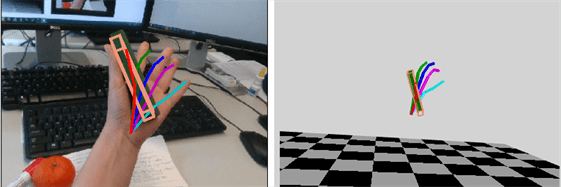} \\ \vspace{7mm}
			\includegraphics[width=0.5\linewidth]{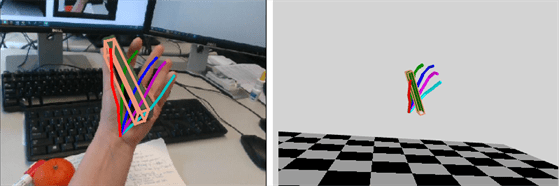}
			&\includegraphics[width=0.5\linewidth]{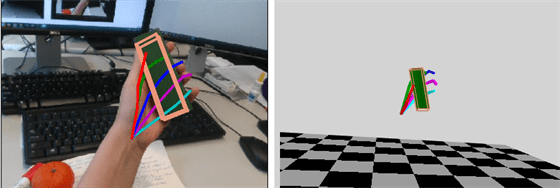} \\ \vspace{7mm}
			\includegraphics[width=0.5\linewidth]{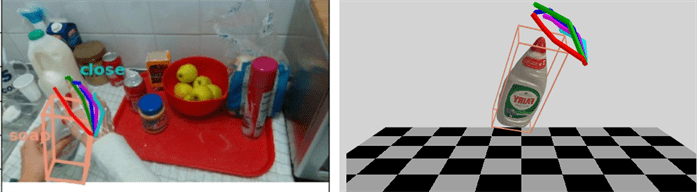}
			&\includegraphics[width=0.5\linewidth]{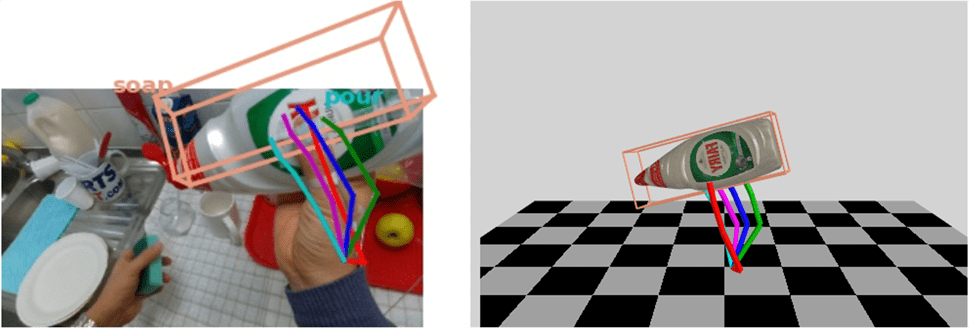} \\	
		\end{tabular}
	} \vspace{0mm}
	\caption{Qualitative  results on the FPHA and EgoDexter dataset. We visualize the 3D hand pose estimates, 3D object bounding boxes which are transformed with the learned 6D object poses, and interaction classes. The proposed approach can handle motion blur, self-occlusions, clutter and complex articulations. We show in the last row failure cases due to an ambiguous action at the frame level (\eg the ``close'' action is predicted, instead of the temporally symmetric ``open'' action) and the occlusion by the viewpoint (\eg the object pose estimate is not very accurate as most of the object is out of the field of view). Best viewed in color.}
	\label{fig:suppresults}
\end{figure*}

\paragraph{Qualitative Results.} We show qualitative results of approach along with some failure cases on the datasets we evaluate our method on in Fig.~\ref{fig:suppresults}. These examples show that our method is robust to severe occlusions, rotational ambiguities in appearance, reflections, viewpoint changes and scene clutter. We provide additional qualitative results in the accompanying video and demonstrate that our \emph{Hand + Object} approach also results in temporally coherent estimates.

\begin{figure*}[t]
	\centering
	\scalebox{0.9}{
		\begin{tabular}{c}
			\includegraphics[width=1\linewidth]{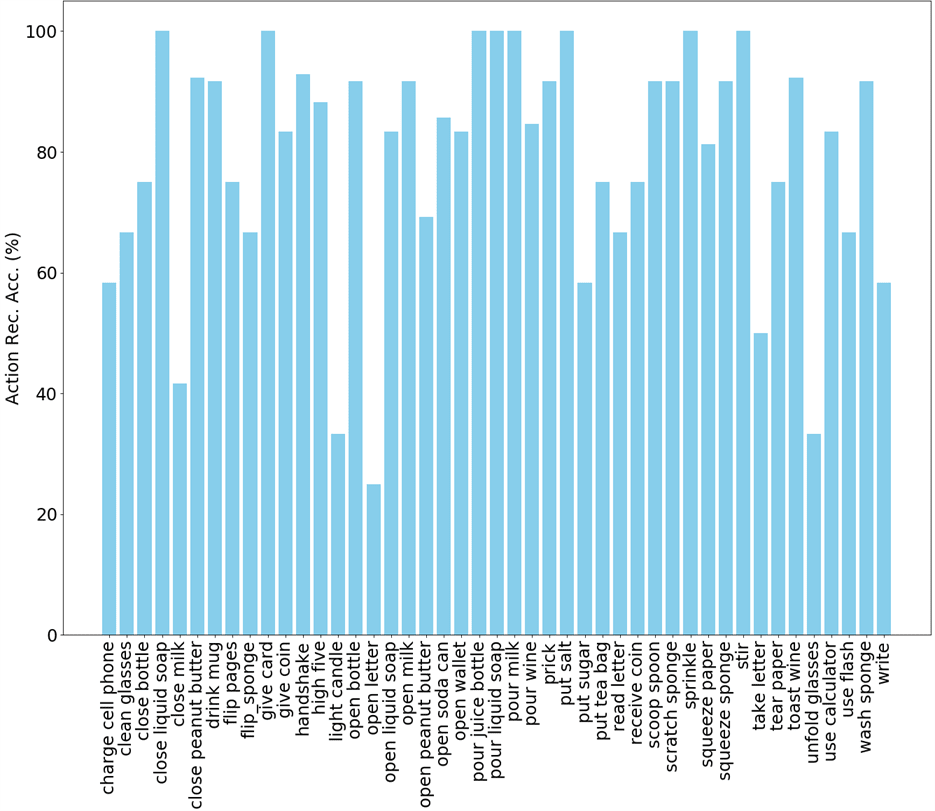}
		\end{tabular}
	} \vspace{0mm}
	\caption{Action-specific recognition accuracies of our approach on the FPHA dataset.}
	\label{fig:accperaction}
\end{figure*}

\paragraph{Recognition Accuracies Per Action.} In Figure~\ref{fig:accperaction}, we show action-specific recognition accuracies on the FPHA dataset. 
While some actions such as `sprinkle', `give coin' and ‘pour juice’ are easily identifiable, actions such as `open letter' and ‘light candle’ are commonly confused, likely because hand poses are more subtle and dissimilar across different trials.

\end{document}